\documentclass[10pt,twocolumn,letterpaper]{article}

\usepackage{wacv}
\usepackage{times}
\usepackage{graphicx}
\usepackage{amsmath}
\usepackage{amssymb}
\usepackage{multirow}
\usepackage{multirow}
\usepackage{makecell}
\usepackage{adjustbox}
\usepackage{subfig}

\graphicspath{ {img/} }

\wacvfinalcopy % *** Uncomment this line for the final submission

\def\wacvPaperID{28} % *** Enter the IJCB Paper ID here

\ifwacvfinal\fi
\begin{document}

%%%%%%%%% TITLE
\title{Predicting Gender From Iris Texture May Be Harder Than It Seems}

\author{Andrey Kuehlkamp\\
University of Notre Dame\\
Notre Dame - IN\\
{\tt\small akuehlka@nd.edu}
\and
Kevin Bowyer
}

\maketitle
\thispagestyle{empty}

%%%%%%%%% ABSTRACT
\begin{abstract}
   Predicting gender from iris images has been reported by several researchers as an application of machine learning in biometrics. Recent works on this topic have suggested that the preponderance of the gender cues is located in the periocular region rather than in the iris texture itself. This paper focuses on teasing out whether the information for gender prediction is in the texture of the iris stroma, the periocular region, or both. We present a larger dataset for gender from iris, and evaluate gender prediction accuracy using linear SVM and CNN, comparing hand-crafted and deep features. We use probabilistic occlusion masking to gain insight on the problem. Results suggest the discriminative power of the iris texture for gender is weaker than previously thought, and that the gender-related information is primarily in the periocular region.
\end{abstract}

%%%%%%%%% BODY TEXT
%------------------------------------------------------------------------
\section{Introduction}

The process of capture, extraction and analysis of information that is complementary to a certain primary biometric treat is called \textit{soft biometrics}, and it can be beneficial for recognition. Soft biometric features like weight, height, ethnicity or gender do not carry the same discriminative power as the primary biometric treat, but they are frequently easier to extract and manipulate. In the case of measurable features like weight or height, one can rely on sensors to accurately capture and link them to the subject. Gender and ethnicity however, have a higher degree of subjectivity -- there is no simple, automated way to assess them.

There are mainly two ways a recognition system can benefit from ancillary information: a) by using the soft biometrics as additional trait for identity confirmation, and thus preventing potential false matches; and b) by providing a way to shorten the verification list, therefore speeding the search process --- if we know we are searching for a woman, it is safe to skip all male enrollments. Another possible application is to be able to generate a general description from unrecognized people; e.g., ``middle-aged, male, caucasian''.  Machine-learning based methods to infer gender from face images  
%and ethnicity \cite{gutta1998ethnic} 
were available since the early 1990s \cite{NIPS1990_sexnet}, but trying to do so from iris images was not attempted until 2007 \cite{Thomas2007}.

Since then, several works have tried 
%(see \cite{Kuehlkamp2017} for a list)  
to use iris information to classify gender. These works have used several different methods and protocols, and reported quite different results. In 
% \newtext{\{reference omitted in compliance to the blind review protocol\}}
our previous work \cite{Kuehlkamp2017} 
we addressed some issues that may have been overlooked by earlier works. There, it is shown that the presence of eye cosmetics and the use of less rigorous cross-validation protocols may have resulted in optimistic estimates of classification accuracy. Furthermore, more recent works \cite{Bobeldyk2016, tapia2018gender, bobeldyk2018gender, Manyala2018gender} suggest that the majority of gender cues is not found in the iris, but in the periocular region that surrounds it. In this sense, our work deepens the understanding of the problem because: a) we use a stricter definition for ``iris-only'' images; and b) unlike the aforementioned works, we account for the presence of cosmetics, which is known to cause interference in gender classification.

To improve our ability to determine whether gender cues are in the iris texture or the periocular region, we present a new dataset for gender from iris, called Gender From Iris -- Cosmetics (GFI-C). This dataset is larger than Gender From Iris (GFI), and its images were manually selected based on image quality and presence of eye cosmetics. One important property of the dataset is that for a portion of the female subjects, we have images both with and without eye cosmetics. Thirty random subject disjoint train/test partitions were created, for four different training groups, to be able to characterize the variance in the accuracy estimate that is due to the train/test split. Basic properties of GFI-C dataset are compared to GFI, in order to validate it.

On the new dataset, we perform a comparison between simple hand-crafted features and deep features extracted by VGGNet \cite{VGG}. Both sets of features are classified by linear Support Vector Machines (SVM). We demonstrate that, unless the periocular image is used, there is very little advantage in the use of deep features. These results suggest the iris texture does not play a primary role in the process.

Convolutional Neural Networks (CNNs) represent the current state of the art in many areas of visual recognition. The VGGNet architecture became very popular after the ImageNet ILSVRC-2014 challenge in which it won the Classification+Localization task\footnote{http://www.image-net.org/challenges/LSVRC/2014/results}. One interesting aspect of CNNs is that in general they are able to learn features from the images on which they are trained.  This ability might bring a specific contribution for gender prediction based on the iris texture, since it has not been demonstrated which particular types of iris features are useful for this purpose. We fine-tune VGGNet-16  for the task of gender prediction, with the intent of investigating if a non-linear classifier can offer an advantage over other approaches. 

\begin{figure}[tb]
    \subfloat[\label{fig:occ_ex00}]{%
        \includegraphics[width=\linewidth,
        trim=4cm 0.7cm 4cm 0.7cm,clip]{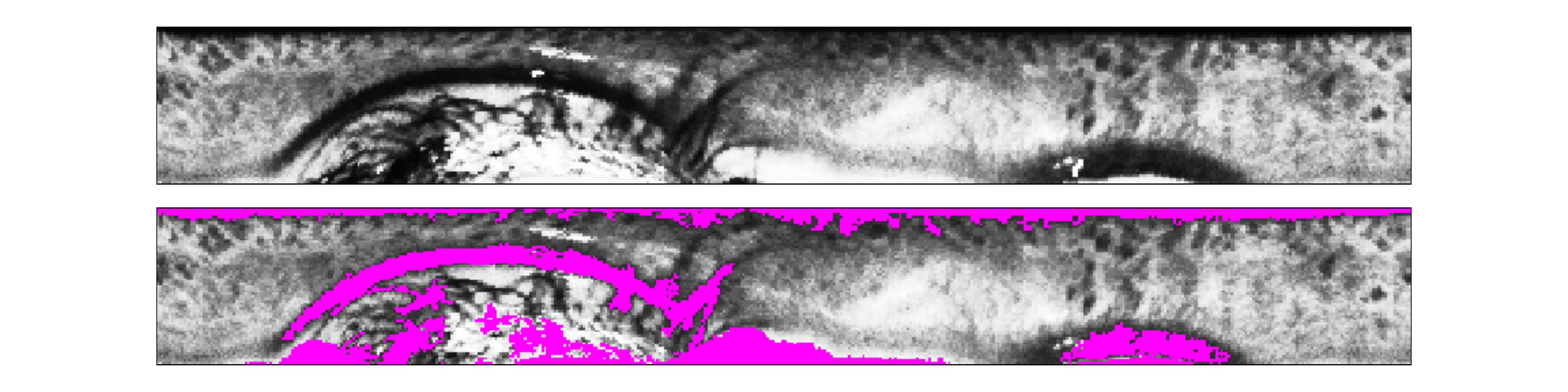}
    }
    \\
    \subfloat[\label{fig:occ_ex01}]{%
        \includegraphics[width=\linewidth,
        trim=4cm 0.7cm 4cm 0.7cm,clip]{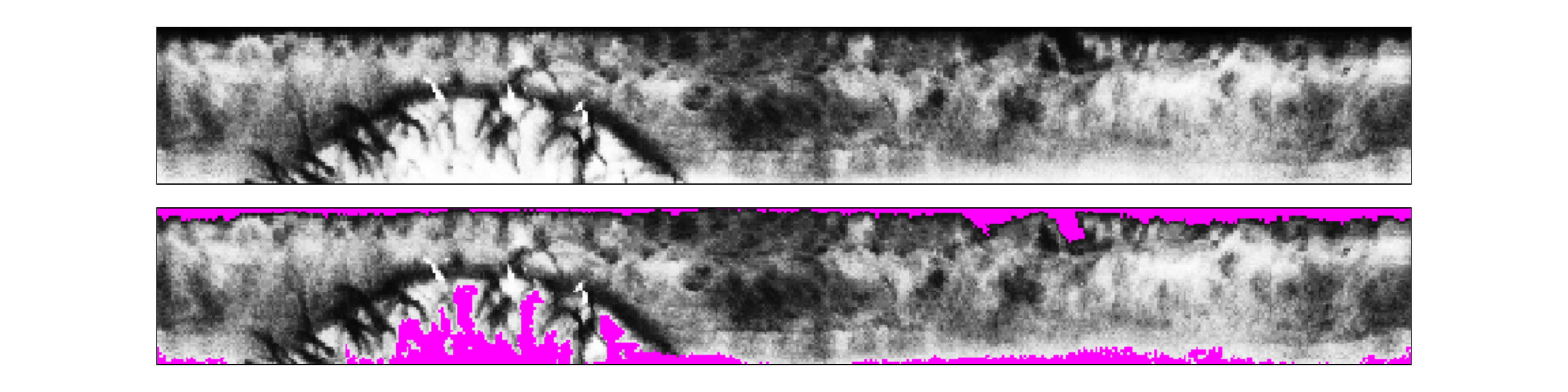}
    }
    \caption{Normalized irises with and without occlusion masking. These examples illustrate how imprecise occlusion masking by the segmenter is confused with iris stroma. These regions can often be affected by eye cosmetics and interfere in gender prediction.}
    \label{fig:occ_examples}
\end{figure}

Considering the evidence suggesting the iris does not contain the majority of gender cues, it is necessary to be careful when performing classification on normalized images. Residual information from outside the iris may be influencing prediction, leading to premature conclusions on the potential of iris texture for gender prediction. Figure \ref{fig:occ_examples} illustrates how the normalized irises can still contain information from outside, even after applying the occlusion mask (magenta regions in the bottom images in \ref{fig:occ_ex00} and \ref{fig:occ_ex01}). 

Recent advances in machine learning systems have focused on performance, but they sometimes lack the ability to be decomposed into understandable components. As a consequence, when such systems fail, it is not uncommon to leave users without a coherent explanation as to what was the reason for failure, as pointed out in \cite{selvaraju2016grad}. In addition to effectiveness, intelligent systems should account for explainability and auditability. In this sense, we want to be able to understand the nature of the cues used in prediction.

To learn more about the influence of occlusions in classification of normalized images, we use an occlusion probability map to eliminate these regions from the images. This allows us to evaluate the discriminative potential of different regions of the iris stroma, revealing interesting implications.

The main contributions of this work are a) a new dataset for the gender prediction problem, comprising 30 random, class-balanced and person-disjoint train/test splits; b) extensive benchmarking and evaluation of the gender prediction problem; c) use of probabilistic masking to investigate the locality of gender cues. The remainder of this paper is organized as follows: in Section \ref{sec:relatedwork} we discuss the progress made in pertinent works; Section \ref{sec:ds_methods} describes the new dataset and the methodological protocol for our experiments; Section \ref{sec:results} presents the classification results, and in Section \ref{sec:loc_cues} presents experiments that try to clarify the locality of gender cues. Finally, in  Section \ref{sec:conclusions} we lay our conclusions.

Some authors prefer the use of the term ``sex'' instead of ``gender''. While the former is biologically determined, the definition of the latter may involve social, personal or psychological factors. However, throughout this work we adopt the term ``gender'' in order to be consistent with the previous work in this area.

%------------------------------------------------------------------------
\section{Related Works}
\label{sec:relatedwork}

A person's gender is a soft biometric attribute that, in association with an individual, can be useful for recognition of that particular individual. Gender prediction based on characteristics of the iris is an idea that has been tried during the last decade with results varying from random accuracy up to $97\%$~\cite{Thomas2007,Lagree2011PredictingEA,Bansal2012,Tapia2014,Fairhurst2015,Tapia2016GenderCF,Bobeldyk2016,Kuehlkamp2017,singh2017gender,tapia2018gender,bobeldyk2018gender,Manyala2018gender}. Most of the works used Support Vector Machines (SVM) as classifiers, with different approaches for feature extraction.

In what was probably the earliest attempt to predict gender from iris texture, \cite{Thomas2007} achieved $75\%$ accuracy with the use of decision trees and hand-crafted geometric features from a dataset of more than 57,000 images.
Using a smaller dataset, but with a stricter evaluation protocol, including person-disjoint train and test partitions, \cite{Lagree2011PredictingEA} reported much more modest accuracy rates, ranging from $47\%$ to $62\%$. They used hand-crafted features and an SVM classifier.

Another work that employed SVM with hand-crafted features \cite{Bansal2012}, seems to have obtained better performance, reporting $83\%$ accuracy. 
A similar result was presented by \cite{Tapia2014}, in the work that originated the Gender-From-Iris dataset. They achieved accuracy of up to $97\%$ using Local Binary Patterns (LBP) features and a SVM. Apart from the higher accuracies, these two works have something else in common: the training and testing procedure was not performed on disjoint partitions. This is important because the classifiers may learn features that characterize a specific person, instead of its gender.

In \cite{Tapia2016GenderCF}, SVM classification using the IrisCode as features was reported to achieve $85\%$ accuracy, on disjoint train and test partitions. On the other hand, \cite{Kuehlkamp2017} takes a more attentive look at evaluation methodology, and finds out that reporting accuracy from a single random train/test split may give rise to biased results. Their accuracy, averaged over 10 repetitions on random disjoint train/test partitions was $66\%$. 

In a similar approach, \cite{Bobeldyk2016} used SVM classification of Binarized Statistical Image Features (BSIF) to explore gender prediction not only from normalized irises, but also from the periocular area. They show that the worst accuracy was obtained from normalized iris images ($65\%$), while the best performance was using the entire eye image ($84\%$). 

More recently, in \cite{bobeldyk2018gender}, the same authors used a similar approach, performing cross-dataset and cross-ethnicity evaluations on gender and ethnicity prediction. Reported accuracy for the ``Iris-Only'' region is about $89\%$. However, the cropped image that the authors refer to as iris-Only,  in fact contains some iris, pupil, sclera, eyelids and eyelashes.

Diverging from the SVM approaches, \cite{singh2017gender} proposed the use of a supervised autoencoder for feature extraction, and later gender and ethnicity prediction. In this work, they introduce the encoding of class label vectors in the autoencoder training (hence the supervision). Gender prediction based on periocular images was performed on person-disjoint train/test splits, and the reported accuracies range from 80 -- 83\%.

In another recent work that is based in neural networks and periocular images, \cite{tapia2018gender} use small CNNs based on the LeNet-5 model, trained individually for each eye. To create a final model, they merge the left and right models and fine tune it to perform gender prediction. Data augmentation techniques were used to train the networks from scratch on the GFI dataset, using a single person-disjoint data split. The final model achieved an accuracy of 87.2\%, an improvement of $\sim 3\%$ over the individual left/right networks.

Similarly, \cite{Manyala2018gender} used a deeper CNN to predict gender from the periocular region extracted from larger, near infrared face images. Using three small datasets, they report accuracies in the range of 86--97\%. This work introduces an interesting variation of experiments, though: removing the eyebrows from the periocular image limited their accuracy to 80--84\%. At the same time, performing classification exclusively on the eyebrow images yielded accuracies from 80--95\%. Although this work seems to have been conducted independently and not focused on iris, its results seem to confirm findings of \cite{Bobeldyk2016, bobeldyk2018gender} and \cite{Kuehlkamp2017}, suggesting a majority of gender cues could be located around the eye.

It is worth noting that most of the works so far do not try to provide an explanation as to what are the kinds of gender cues that are being used in classification, or their specific localization. In this work, we present a new dataset for gender prediction from iris, trying to better analyze the distortions caused by the use of eye cosmetics and iris occlusions. One of the features of this dataset is that for a subset of the female subjects, images with and without eye cosmetics are available. We compare properties of this dataset with the previous one for validation. We explore the use of data-driven features extracted using VGGNet \cite{VGG} convolutional network, and compare results with hand crafted features.
In addition, we investigate the locality of learned features using probabilistic occlusion masking of the images.

%------------------------------------------------------------------------
\section{Dataset and Methods}
\label{sec:ds_methods}

The Gender From Iris (GFI) dataset \cite{Tapia2014} was the first dataset created specifically for the problem of determining the gender of an individual based on the iris texture. It was later updated \cite{Tapia2016GenderCF} in order to be completely person-disjoint (i.e., one image per eye for each subject).

Previous work \cite{Kuehlkamp2017} reported a significant interference in the classification potentially caused by the presence of eye cosmetics in the female population. The GFI dataset was not collected having in mind potential distortions caused by the use of cosmetics, and yet approximately $60\%$ of the female subjects wear some kind of eye makeup.
To address this issue more carefully, we selected a new dataset from available images in the 
% [\textit{\color{blue}text was modified to ensure review anonymity}].
Biometrics Research Grid of the University of Notre Dame. 
All images were captured using a LG-4000 sensor, between the years of 2008 and 2016. 
Other aspects were taken into consideration while manually selecting the images: The annotators tried to select the best images of each subject in terms of alignment, focus, illumination and occlusion.

The new dataset, called GFI-C, is composed of 6240 images of 2005 subjects. Again, approximately 60\% of the female images have some type of cosmetics, while none of the male images have cosmetics. 
For each eye of every female subject, annotators tried to find one image with cosmetics and one without.
Only 143 female subjects fall into this case, while most subjects have either images with or without cosmetics (as shown in Tables \ref{tab:gfic_subject_compos} and \ref{tab:gfic_img_compos}).
For male subjects, two images of each eye were selected to compose the dataset. 
 
\begin{table}[htb]
    \centering
    \small
    \caption{Composition of GFI-C by subjects}
    \label{tab:gfic_subject_compos}
    \begin{tabular}{c|c|c|c}
        \hline
        Gender                  & Cosmetics & \# Subjects & \%     \\ \hline
        \multirow{3}{*}{Female} & No        & 318         & 15.9 \\
                                & Yes       & 562         & 28.0 \\
                                & Both      & 143         & 7.1  \\ \hline
        Male                    & No        & 982         & 49.0 \\ \hline
    \end{tabular}
\end{table}

\begin{table}[htb]
    \centering
    \small
    \caption{Composition of GFI-C by images}
    \label{tab:gfic_img_compos}
    \begin{tabular}{c|c|c|c}
        \hline
        Gender                  & Cosmetics & \# Images & \%     \\ \hline
        \multirow{2}{*}{Female} & No        & 914       & 14.6 \\
                                & Yes       & 1402      & 22.5 \\ \hline
        Male                    & No        & 3924      & 62.9 \\ \hline
    \end{tabular}
\end{table}

% % this figure is for the next section
% \begin{figure*}[!tb]
%     \centering
%     \includegraphics[width=\linewidth,
%         trim=0.5cm 1cm 0.5cm 1cm]{ds_comparison.pdf}
%     \caption{A comparison of several aspects of normalized irises from GFI and GFI-C datasets.}
%     \label{fig:ds_comp}
% \end{figure*}

To reduce the chance of biased partitioning (as described in \cite{Kuehlkamp2017}), thirty random train/test splits were created, using approximately 80\% of the images for training and 20\% for testing. The exact size of the partitions may vary depending on the number of images available for each subject. However, splits are formed by sampling from the dataset, to ensure the balance between male and female images in each partition.
All the splits are person-disjoint, i.e. the subjects that are used in the training partition are not present in the testing partition.
Also, the number of subjects in each group (Males, Females With Cosmetics and Females No Cosmetics) was balanced, to avoid any type of biasing in the process.
Finally, permutations were generated in four different versions, to evaluate predictions with different training groups: 
\begin{enumerate}
    \vspace{-3mm}
    \item \textit{All Subjects} -- training partition contains samples from all groups: Males, Females No Cosmetics and Females With Cosmetics;
    \vspace{-3mm}
    \item \textit{FNC} -- training partition contains only Males and Females No Cosmetics;
    \vspace{-3mm}
    \item \textit{FWC} -- training partition contains only Males and Females With Cosmetics;
    \vspace{-3mm}
    \item \textit{Paired} -- training partition is composed by pairs of images: each female have one image with cosmetics and one without, while each male has two images (without cosmetics).
\end{enumerate}
\vspace{-3mm}
In all cases, the testing partition contains samples from all groups.

In order to assess the soundness the new GFI-C dataset, we performed an analysis of aspects that may cause interference in gender prediction, and compared the results with the previous GFI dataset.

Intensity is the primary information in an image. It has been shown that by simply establishing a threshold for the average intensity it is possible to predict gender with approximately $60\%$ accuracy \cite{Kuehlkamp2017}. We compared the mean intensity of the entire normalized iris, and also the mean intensity in each of five horizontal bands, starting from the pupil-iris boundary until the iris-sclera border.

Iris occlusion is another important aspect, in the sense that it determines how much of the actual iris texture is visible in the image. In the segmentation/normalization process, an occlusion mask is created to exclude objects that obstruct the iris texture, typically eyelids and eyelashes. Similar to occlusion, pupil dilation is another factor that determines the visible pupil area in the image. A highly dilated pupil compresses the stroma to a narrow area and highly constricted pupil stretches it, causing distortions to its natural relaxed state. It has already been shown that pupil dilation degrades the accuracy of iris recognition \cite{hollingsworth2009}. Image sharpness could also cause local intensity distortions in the iris texture and thus disturb classification. We performed a sharpness assessment of the images in both datasets following the ISO/IEC 29794-6:2014(E) \cite{ISO29794-6}. 

Comparison shows that despite having been manually selected and being significantly larger, GFI-C still holds the same basic properties of the previous dataset. A similar proportion of the female population ($\sim60\%$) contains some type of eye cosmetics. Dilation and ISO Focus are significantly lower in the new dataset. These can be attributed to the manual selection process, because annotators intentionally favored images with lower pupil dilations and sharper focus. Finally, the imbalance between the gross number of male and female images is compensated in the formation of the train/test splits.

%------------------------------------------------------------------------
\section{Classification Results}
\label{sec:results}

\subsection{Hand Crafted features with SVM classifier}

% \begin{figure*}[tbp]
%     \centering
%     \includegraphics[width=1\textwidth,
%         trim=0.5cm 1cm 0.5cm 1cm]{hcf_distributions.pdf}
%     \caption{Analysis of hand crafted features potential for gender classification. Discrimination power for each individual feature is marginal. In several cases, the potential to discriminate between \textit{With Cosmetics}/\textit{No Cosmetics} is higher than gender.}
%     \label{fig:hcf_dist}
% \end{figure*}

Upon normalization of the images, 18 different simple features related to the iris location and intensity were extracted. These features comprehend $xy$ coordinates, radius and area for the iris and the pupil, ISO focus, occlusion and dilation ratios, as well as average intensity in the overall iris and in five bands going from the pupil until the sclera boundary. We grouped distributions by different properties (gender and presence of cosmetics), and calculated the Equal Error Rate (EER) in order to assess their potential for classification. An EER of 50\% means the classification power of that specific feature is equivalent to random chance. Since we are dealing with binary classification, the larger the EER absolute difference from 50\%, the better its discriminative power. A graph of these distributions is provided in the supplementary material to this paper. This assessment revealed that the discriminative power of any of these features is not higher than 6.84\% for gender and 10.03\% for presence of cosmetics (Iris Y coordinate). This supports the findings of \cite{Kuehlkamp2017}, in the sense that it is not possible to know if the classifier is actually discriminating gender or cosmetics.

\begin{figure*}[tbp]
    \centering
    \includegraphics[width=1\textwidth,
        trim=0.5cm 0.7cm 0.5cm 0.5cm, clip]{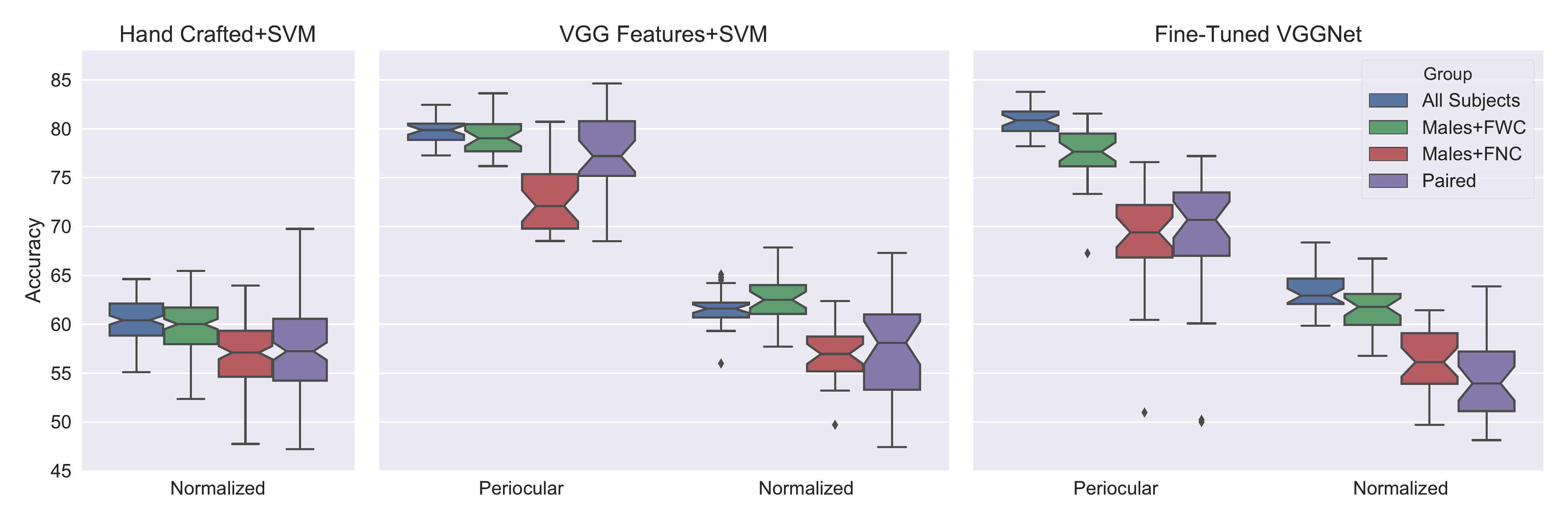}
    \caption{Gender classification accuracy using: 1) Hand Crafted Features and SVM; 2) VGG deep features and SVM; and 3) a fine-tuned version of VGGNet \cite{VGG}. Dots represent outlier results.}
    \label{fig:overall_acc}
\end{figure*}

Considering it is possible that the combination of features in a multidimensional hyperplane could allow better separation for gender, these features were fed into a linear SVM. The mean accuracy across 30 train/test partitions was $61.23\% \pm1.58$. To isolate possible effects of eye cosmetics, additional experiments were performed in different training groups: FWC, FNC and Paired. The left chart on Figure \ref{fig:overall_acc} shows the result of these experiments.

Training on FWC reflects on a slightly lower average accuracy and a wider distribution, which could be due to a smaller training set. Training on FNC results in a significant decrease in accuracy, and the distribution spreads towards lower values, showing that information that is important for the decision was lost. Finally, if training is performed on the Paired group, mean accuracy is still low, but the range of accuracy is wider. This suggests that the reintroduction of cosmetics into the training set helps gender classification, even when using a much smaller training set.

\subsection{VGG features with SVM classifier}

At least two disadvantages may be attributed to the use of these simple hand crafted features: first, most of them are not directly relatable to gender attributes, and second, almost all of them are a byproduct of the iris normalization process, which prevents us from performing classification on periocular images. To make use of more comprehensive and potentially more powerful features, we decided to apply deep learning to the feature extraction step. However, to make results directly comparable to our previous results, we use a linear SVM to classify the deep features.
Using VGGNet-16 \cite{VGG} network trained on ImageNet, we extracted 4096 features of the iris images at the first fully connected layer, and classified them using a Linear SVM. Since now features are data driven based on the whole image, we can classify periocular as well as normalized irises. The results of this experiments are shown in the center chart on Figure \ref{fig:overall_acc}.

There is a substantial difference between the accuracy of periocular and normalized images. This difference confirms the findings of \cite{Bobeldyk2016} and \cite{Kuehlkamp2017}, which suggest the majority of the cues used in the prediction comes from the periocular region, instead of the iris texture.
It can also be observed here that the same drop in accuracy happens when cosmetic images are removed from training, regardless of the type of image that is being used (periocular or normalized). This suggests the segmentation process is not being completely effective in removing non-iris components from the normalized images, and that these components continue being used in the classification.

A Kolmogorov-Smirnov test shows the distributions of Hand Crafted and VGG features for normalized images are different at a 5\% significance level, although the means are very similar ($60.4\%\pm2.2$ and $61.6\%\pm1.8$ respectively). On periocular images, the linear SVM with VGG features achieved an average accuracy of $79.8\%\pm1.3$.

\subsection{Fine-tuned VGG}

Additionally to feature extraction, VGGNet-16 was also used to perform classification. Like before, the network was initialized with weights learned from ImageNet. Next, all the convolutional layers were frozen, and training was performed on GFI-C dataset, using the previously defined protocol. These results can be seen in the right chart on Figure \ref{fig:overall_acc}.

\begin{table*}[htb]
\centering
\caption{Comparison of Classification Accuracy (\%) and Standard Deviation between GFI and GFI-C, Using Different Features and Classifiers}
\label{tab:ds_comparison}
\begin{adjustbox}{max width=\textwidth}
\begin{tabular}{c|c|c|c|c|c|c|c}
\hline
                            & \makecell{Features/\\Classifier} & \multicolumn{2}{c|}{\makecell{Hand-crafted/\\Linear SVC}} & \multicolumn{2}{c|}{\makecell{VGG features/\\LinearSVC}} & \multicolumn{2}{c}{\makecell{VGG features/\\VGG}}\\ \cline{2-8}
                            & Dataset             & GFI                       & GFI-C                     & GFI                       & GFI-C                     & GFI                       & GFI-C                     \\ \hline
\multirow{3}{*}{Normalized} & All Subjects        & 58.4$\pm$2.4 & 61.2$\pm$1.6 & 60.0$\pm$1.8   & 61.6$\pm$1.8 & 60.1$\pm$3.0 & 63.4$\pm$1.9 \\
                            & Males+FNC           & 54.2$\pm$3.0 & 58.1$\pm$3.4 & 55.1$\pm$3.1 & 57.1$\pm$2.9 & 54.9$\pm$2.8 & 56.3$\pm$3.4 \\
                            & Males+FWC           & 58.0$\pm$2.7 & 59.9$\pm$2.6 & 61.0$\pm$2.0   & 62.5$\pm$2.3 & 61.2$\pm$2.5 & 61.6$\pm$2.4 \\ \hline
\multirow{3}{*}{Periocular} & All Subjects        & - & - & 80.5$\pm$1.6 & 79.8$\pm$1.3 & 71.5$\pm$8.2 & 80.8$\pm$1.3 \\
                            & Males+FNC           & - & - & 75.1$\pm$3.5 & 72.8$\pm$3.3 & 56.3$\pm$5.5 & 69.0$\pm$5.2   \\
                            & Males+FWC           & - & - & 79.5$\pm$2.4 & 79.2$\pm$2.0 & 73.3$\pm$7.4 & 77.5$\pm$2.9 \\ \hline
\end{tabular}
\end{adjustbox}
\end{table*}

\begin{figure}[!htb]
    \centering
    \begin{tabular}{@{}c@{}c@{}}
        % \subfloat[\label{fig:roc_gfi_norm}]{%
        %     \includegraphics[width=0.47\linewidth,
        %      trim=0.2cm 0.9cm 0.5cm 0.9cm, clip]{roc_gfi_norm}
        % } &
        \subfloat[\label{fig:roc_gfic_norm}]{%
            \includegraphics[width=0.52\linewidth,
            trim=0.2cm 0.55cm 0.5cm 0.9cm, clip]{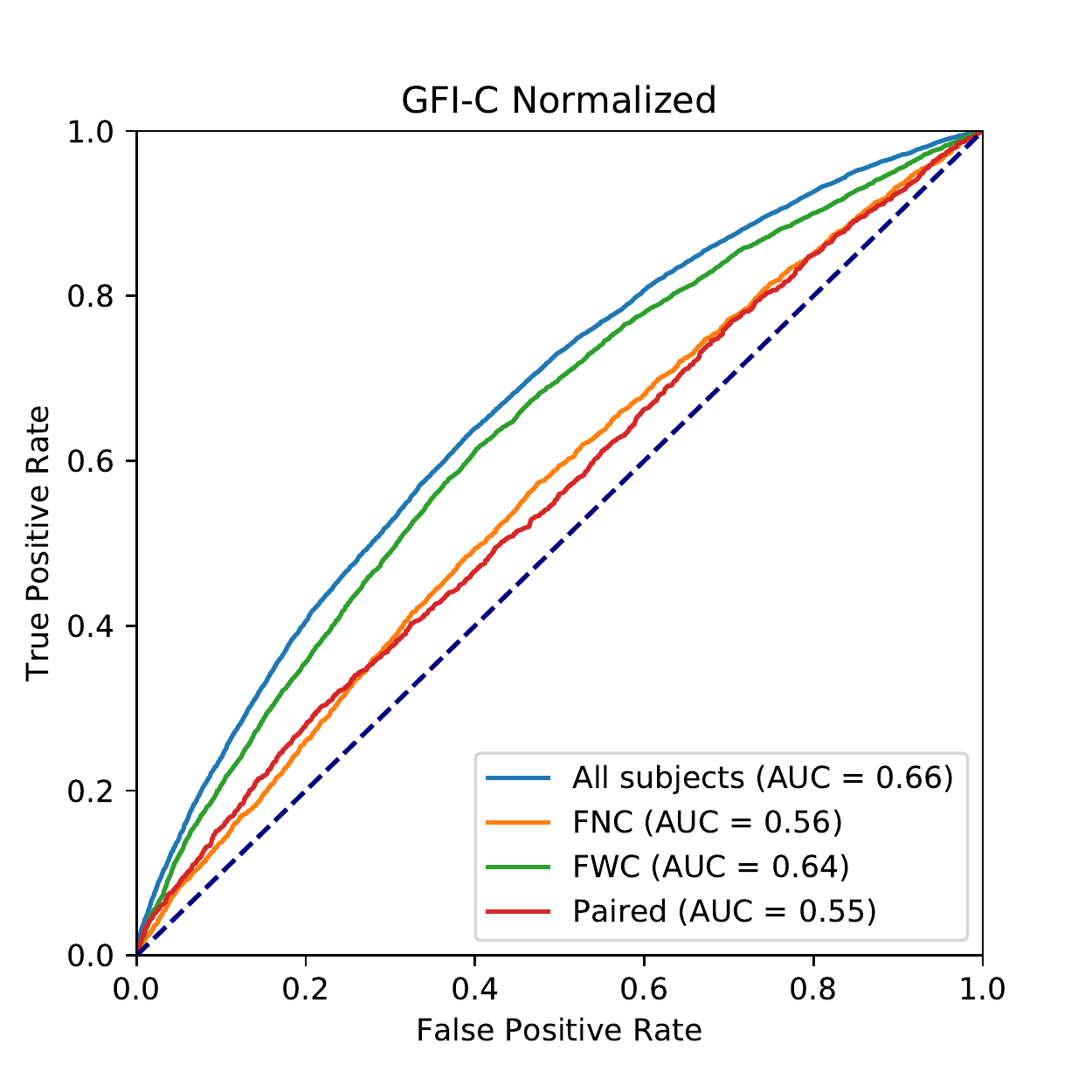}
        } &
        % \\ \rule{0pt}{-10ex}
        % \subfloat[\label{fig:roc_gfi_peri}]{%
        %     \includegraphics[width=0.47\linewidth,
        %      trim=0.2cm 0.55cm 0.5cm 0.9cm, clip]{roc_gfi_peri}
        % } &
        \subfloat[\label{fig:roc_gfic_peri}]{%
            \includegraphics[width=0.5\linewidth,
            trim=0.7cm 0.55cm 0.5cm 0.9cm, clip]{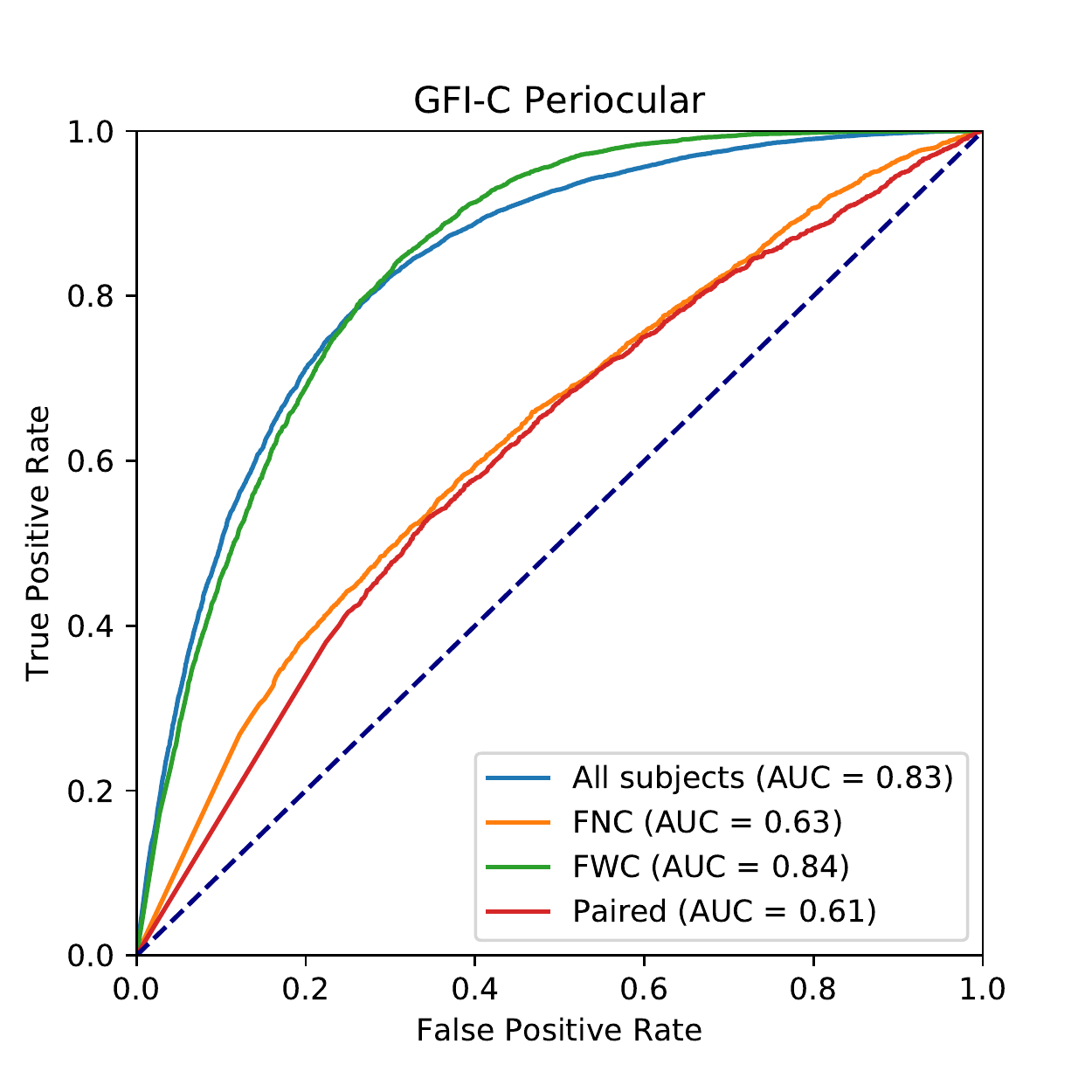}
        }    \end{tabular}
    \vspace{-5mm}
    \caption{ROC curves for gender classification on different training groups using VGGNet.}
    \label{fig:rocs}
\end{figure}

Once again, important accuracy reductions happen when the training set does not contain cosmetics. However, in this case the Paired training group had an accuracy distribution more similar to FNC, unlike what happened with SVM classifiers. As it could be expected, the fine-tuned version of VGGNet performs significantly better than a linear SVM on VGG features, both in periocular and normalized images. The mean accuracies were $80.8\%\pm1.3$ and $63.4\%\pm1.8$, respectively. The same kind of trend is also found in the ROC curves and their respective AUCs, as shown in Figure \ref{fig:rocs}: Accuracy is significantly hampered if the training data does not contain cosmetics.

\subsection{Discussion}

It is possible to perform gender classification on NIR ocular images with different degrees of success, depending on the input: periocular or normalized iris. Periocular images are clearly better than normalized irises. Complex non-linear classifiers (VGGNet) yield better results over simpler classifiers (linear SVM), even when the latter is fed with deep features.

As it was previously suggested 
% \newtext{\{reference omitted in compliance to the blind review protocol\}}
\cite{Kuehlkamp2017}
, eye cosmetics play a significant part in gender prediction from iris or periocular images. Experiments with hand crafted and VGG features show that eye cosmetics make gender classifiers more efficient. This is potentially caused by the introduction of distinctive contrast patterns in regions adjacent to the iris. The effect is still present in normalized iris classification, because these regions are not completely eliminated by the segmentation process. A comparison of the classification results with the previous dataset is presented in Table \ref{tab:ds_comparison}. The same type of trend across training groups happen in both datasets, supporting the hypothesis that cosmetics are influencing gender classification.

Perhaps the most thought-provoking result is that using a small (and weak) set of hand crafted features based on the normalized iris image, together with a simple linear SVM, it is still possible to deliver accuracies that are in the same range of VGGNet. Strictly speaking, an accuracy in the range of 60 -- 65\% for a binary classification problem is not a good result: it is only 15\% better than random guessing. Considering still a) the bias that is introduced by the consistent use of cosmetics on the female population; and b) the occlusion regions that cannot be easily eliminated from the normalized iris, there is very little that remains as the potential role of the iris stroma in gender prediction.

In this sense, we devised a set of experiments to get some insight as to the location and types of features that are used for gender classification in VGGNet.

%------------------------------------------------------------------------
\section{Localization of Gender Cues}
\label{sec:loc_cues}

Despite the variety of works in gender from iris, no good explanation was provided about the origin of the gender cues used in the process. Recent papers have determined that the periocular region is richer than the iris  in gender cues, but they fall short of providing a more precise spatial localization for them. In this section, we describe how probability occlusion masks were used to investigate the location of gender cues.

\begin{figure}[!b]
    \centering
    \includegraphics[width=\linewidth,
     trim=0cm 0cm 9.5cm 0.5cm,clip]{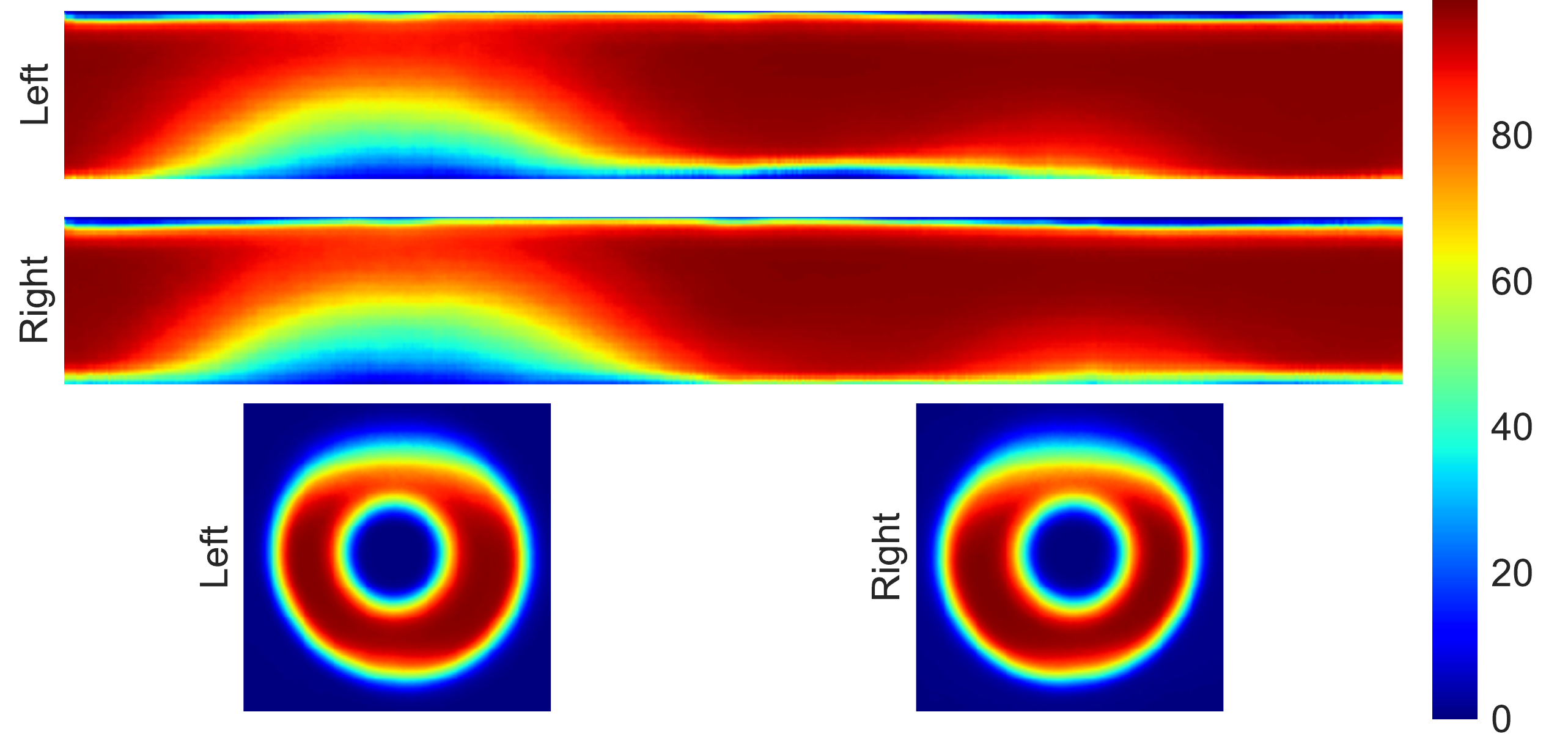}
    \caption{Non-Occlusion probability for normalized and cropped irises.}
    \label{fig:nop}
\end{figure}

\begin{figure*}[!htb]
    \centering
    \begin{tabular}{@{}c@{}}
        \subfloat[\label{fig:pmask_norm_acc}]{%
            \includegraphics[width=\textwidth,
             trim=0.5cm 1.1cm 0.5cm 0.8cm]{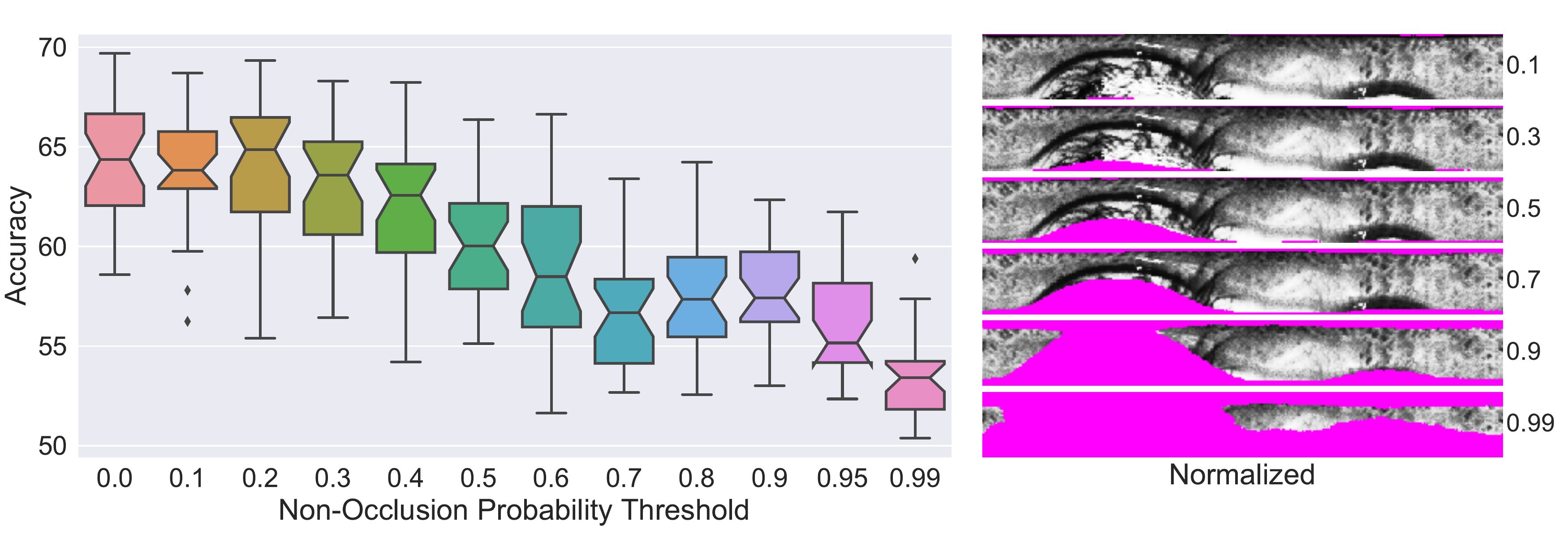}
        } \\
        \subfloat[\label{fig:pmask_peri_acc}]{%
            \includegraphics[width=\textwidth,
            trim=0cm 0cm 1cm 0.7cm,clip]{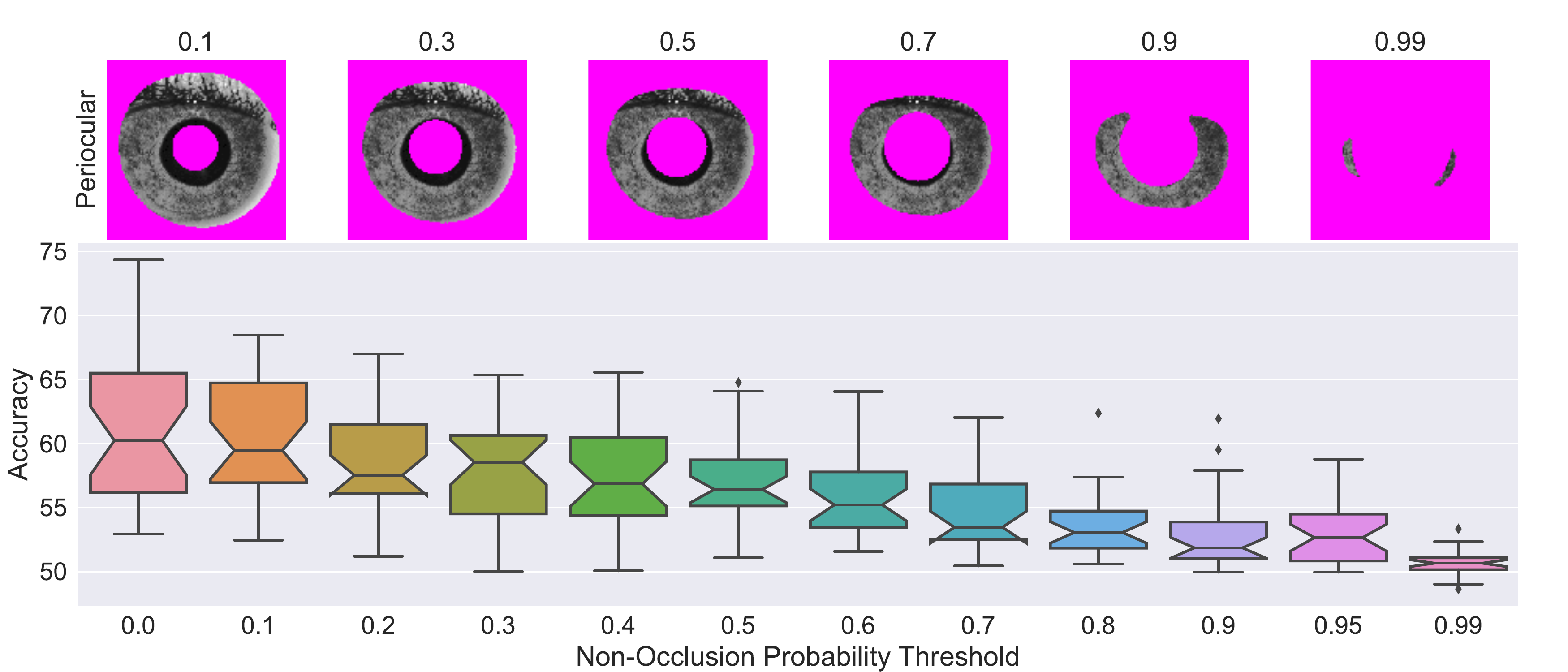}
        }
    \end{tabular}
    \caption{Gender classification accuracy using probabilistic masking in normalized (a) and periocular (b) images. Dots in the box plots represent outlier results.}
\end{figure*}

Using the occlusion mask that is generated for each iris upon segmentation/normalization, we calculated non-occlusion probability maps for left and right eyes, for normalized images and also for the cropped irises. These maps (Figure \ref{fig:nop}) show the probability that a certain pixel will not be occluded by something like eyelids or eyelashes. In this specific case, it defines the probability that a pixel of the region belongs to the iris stroma texture or something else.

Using the non-occlusion probability maps at different thresholds, the irises were masked, and then used to train VGGNet-like classifiers. Images resulting from the use of these probabilistic masks can be seen on the right side of Figure \ref{fig:pmask_norm_acc} for normalized irises, and on the top portion of Figure \ref{fig:pmask_peri_acc} for periocular images. In both types of images, even at thresholds between 50--70\% it is still possible to identify eyelash and eyelid regions in the image. 

Using a probability mask like this serves two purposes: first, to hide occlusions that can be learned by the classifier as gender cues; and second, to provide a single occlusion mask that is used for all images, preventing that each image individual occlusion mask end up being learned as gender cues. In this way, we tried to ensure that gender features are being learned from the iris stroma, and not from residual features of other regions, or artifacts (occlusion masks).

Classification experiments were performed on periocular and normalized images, using probabilistic masks thresholded from 0 to 0.99. Results of this classification on normalized images are presented in Figure \ref{fig:pmask_norm_acc}. They suggest that gender prediction accuracy is significantly reduced as occlusion-prone regions are suppressed from the images. One could argue against this approach by claiming that the decreasing trend in accuracy is caused merely because information decreases when the non-occlusion probability threshold is incremented.

Accuracy results on normalized images not using any masking (threshold 0) are similar to previous literature, with an average of $64.2 \pm 2\%$. When pixels that have an occlusion probability of more than 70\% are masked, mean accuracy drops to $56.8 \pm 3\%$. Finally, when only pixels that have more than 99\% probability of non-occlusion are used, the mean accuracy is only $53.4 \pm 2\%$.

Classification accuracies for periocular images with probabilistic masks was similar those of normalized images, but with noticeably lower values. A possible explanation for this is because normalization ensures the iris localization, while the probability mask focuses on where \textit{most}
 irises are located. This could cause problems for cases where the iris is not centered in the image. Nevertheless, we believe this approach to be a more adequate implementation of ``iris-only'' classification than those  presented on \cite{Bobeldyk2016} and \cite{bobeldyk2018gender}

\begin{figure}[!ht]
    \begin{tabular}{@{}c@{}}
         \subfloat[Iris, non-occlusion probability $>$ 70\%\label{fig:iris70}]{%
            \includegraphics[width=0.43\linewidth,
             trim=10cm 9cm 0.5cm 0.5cm,clip]{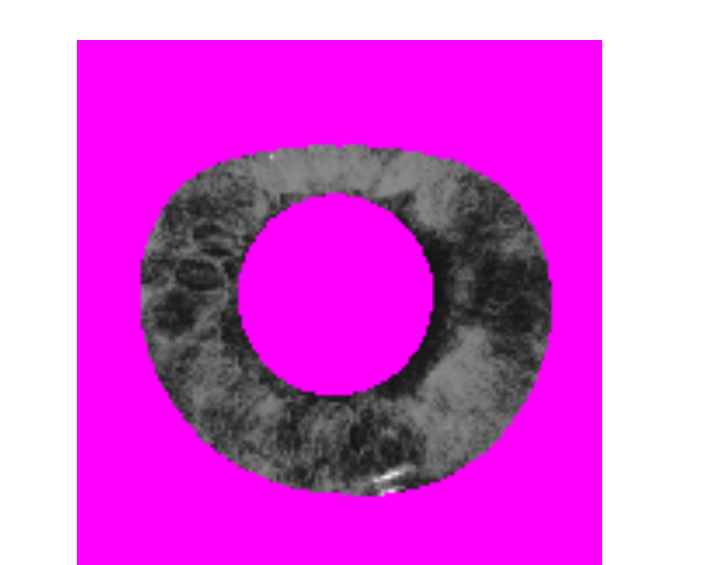}}
             \\
         \subfloat[Non-iris, area equivalent to non-occlusion probability $>$ 70\%\label{fig:non-iris70}]{%
            \includegraphics[width=0.43\linewidth,
             trim=10cm 1.7cm 0.5cm 8cm,clip]{iris-non-iris_comparison}}
    \end{tabular}
    \hfill
    \begin{tabular}{@{}c@{}}
         \subfloat[Prediction accuracy using different image regions.\label{fig:ini-acc}]{%
            \includegraphics[width=0.5\linewidth,
             trim=0.9cm 0.7cm 8cm 1cm,clip]{iris-non-iris_comparison}}
    \end{tabular}
    \label{fig:iris-non-iris_comp}
    \caption{Gender prediction accuracy is higher when using occlusion-influenced iris area, even when the size of the area is equivalent.}
\end{figure}

As considered earlier, a decreasing useful image area could be the responsible for the decreasing trend in accuracy. To investigate this assumption, we compared prediction accuracy obtained with the iris region with a probability of non-occlusion greater than 70\% (Fig. \ref{fig:iris70}) against accuracy on a non-iris region, but with similar area (Fig. \ref{fig:non-iris70}). The comparison in Figure \ref{fig:ini-acc} shows better results were obtained with the region that is least likely to contain iris texture. Also, the non-iris region yields higher variance, suggesting it is possible to get unusually higher accuracies that are not representative of the actual predictive power. This strongly suggests that there is something near the iris, other than the iris, that is correlated with gender.

This evidence shows that most of the network ability to learn gender cues comes from regions that do not correspond to the iris stroma. In this case, we observe a clear and significant reduction of the performance, as regions subject to external interference are removed from the images. These results reinforce the notion that the iris stroma texture has very low potential for gender prediction.

%------------------------------------------------------------------------
\section{Conclusions}
\label{sec:conclusions}

We presented a new, larger dataset for the Gender From Iris problem. Images were manually selected in favor of good quality and low occlusion, and it contains gender and cosmetic ground truth annotations. In addition to that, the dataset provides 30 random,  person disjoint train/test splits. There are also partitions for specific training groups -- FNC, FWC and Paired. All partitions are balanced for gender and group. Its basic properties are similar to the GFI dataset, but it has more images and subjects with paired samples (with/without cosmetics). 

The discriminative power of different types of features was compared, using linear SVM as a baseline classifier. Gender prediction was evaluated using periocular and normalized images. Also, a deep CNN was fine-tuned for the problem of gender prediction. Our experiments showed that the discrimination potential of individual hand-crafted features is very low, and yet their combination can yield $\sim 60\%$. Furthermore, the gender discriminative potential of individual features is frequently equivalent to the presence of cosmetics, making it hard to ascertain if classification actually comes from one or another. Experiments performed on different training groups show a significant decrease in accuracy when cosmetics are removed from training, confirming the interference of these in the problem.

Since the set of hand-crafted features is composed of general and aggregate information on the segmented iris, it is hardly capable of describing complex textures. However, in a direct comparison, deep features do not offer much improvement, which casts suspicion on whether the deep features are indeed representing something more complex. Predicting gender from periocular images and from iris images are two very distinct problems, and the last is much harder. Regardless of the classifier (SVM or CNN), there is a clear and significant difference between predicting gender from periocular images and normalized iris images -- on average, periocular images are at least 17\% more accurate than normalized images.

Gradually applying probability occlusion masks to normalized images before training shows that the more occlusions are eliminated, the less predictive power remains in the images. Evidence suggests that, as occlusions are eliminated from normalized iris, the remaining gender discriminative potential trends toward random chance. 

%------------------------------------------------------------------------
\section{Acknowledgement}

% \textit{\color{blue}Section omitted to ensure anonymity in the review process.}
This research was partially supported by the Brazilian Coordination for the Improvement of Higher Education Personnel (CAPES)  through  grants  BEX  12976/13-0.

{\small
\bibliographystyle{ieee}
\bibliography{references}
}

\end{document}